\documentclass[runningheads]{llncs}

\usepackage{multirow}
\usepackage{eccv}

\usepackage{eccvabbrv}

\usepackage{graphicx}
\usepackage{booktabs}
\usepackage{capt-of}

\usepackage{hyperref}
\usepackage[capitalize]{cleveref}
\crefname{section}{Sec.}{Secs.}
\Crefname{section}{Section}{Sections}
\crefname{table}{Tab.}{Tabs.}
\Crefname{table}{Table}{Tables}

\usepackage{orcidlink}

\begin{document}

\title{From Drop-off to Recovery: A Mechanistic Analysis of Segmentation in MLLMs} 

\titlerunning{From Drop-off to Recovery: Segmentation in MLLMs}

\author{Boyong Wu\inst{1,2} \and
Sanghwan Kim\inst{1,2,3} \and
Zeynep Akata\inst{1,2,3}}

\authorrunning{B.~Wu et al.}


\institute{\textsuperscript{1}Technical University of Munich \quad
\textsuperscript{2}Helmholtz Munich \\
\textsuperscript{3}Munich Center for Machine Learning (MCML)}

\maketitle

{\centering
\includegraphics[width=\textwidth]{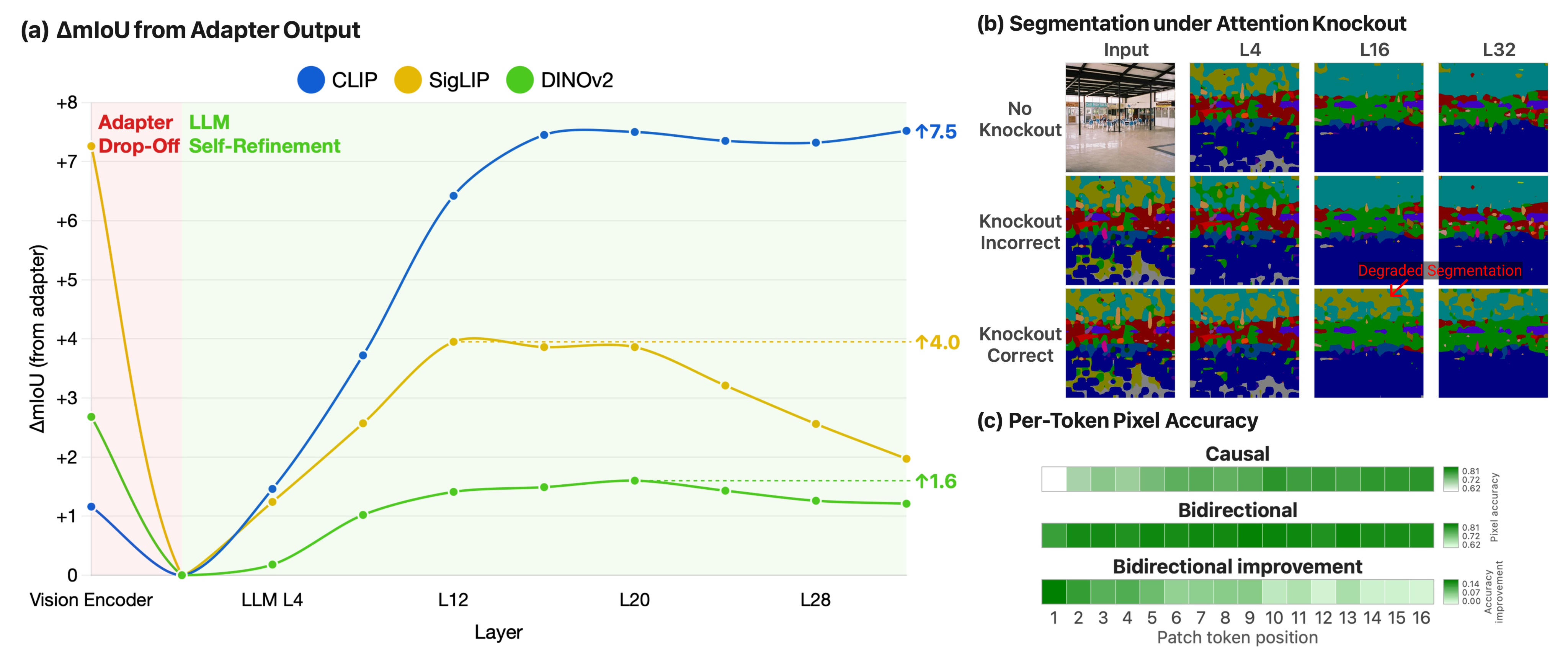}
\captionof{figure}{\textbf{Overview of main findings.}
\textbf{(a)}~Layerwise linear probing on ADE20K: the adapter introduces a
representation drop-off, but LLM layers progressively recover segmentation
quality.
\textbf{(b)}~Attention knockout on conflicting class pairs: knocking out attention from correctly classified tokens degrades segmentation, confirming that
cross-token self-refinement is driven by semantic anchors.
\textbf{(c)}~Per-token pixel accuracy at an intermediate LLM layer: causal attention
starves early position tokens of semantic anchors, while bidirectional attention among
image tokens alleviates this bottleneck.}
\label{fig:main}\par}

\begin{abstract}
Multimodal Large Language Models (MLLMs) are increasingly applied to pixel-level vision tasks, yet their intrinsic capacity for spatial understanding remains poorly understood. We investigate segmentation capacity through a layerwise linear probing evaluation across the entire MLLM pipeline: vision encoder, adapter, and LLM. We further conduct an intervention based attention knockout analysis to test whether cross-token attention progressively refines visual representations, and an evaluation of bidirectional attention among image tokens on spatial consistency. Our analysis reveals that the adapter introduces a segmentation representation drop-off, but LLM layers progressively recover through attention-mediated refinement, where correctly classified tokens steer misclassified neighbors toward the correct label. At early image token positions, this recovery is bounded by causal attention, which bidirectional attention among image tokens alleviates. These findings provide a mechanistic account of how MLLMs process visual information for segmentation, informing the design of future segmentation-capable models.


\end{abstract}
\section{Introduction}
\label{sec:intro}
Detecting and segmenting content in visual scenes is foundational for applications in robotics, AR/VR, autonomous driving, and medical imaging \cite{he_mask_2018,cheng_masked-attention_2022,zhang_survey_2024}. Vision Transformers pretrained with contrastive image-text objectives or self-supervised distillation have emerged as powerful feature extractors: their patch-level embeddings capture rich dense semantics and transfer strongly to segmentation when paired with lightweight decoders or linear probes \cite{banani_probing_2024,radford_learning_2021,oquab_dinov2_2024}. More recently, large-scale pretraining has yielded specialized vision encoders such as SAM \cite{kirillov_segment_2023,ravi_sam_2024,carion_sam_2025} that further push segmentation performance.
In parallel, Multimodal Large Language Models (MLLMs) have demonstrated impressive capabilities in multimodal reasoning, instruction following, and language conditioned control \cite{bai_qwen25-vl_2025,chen_expanding_2025, liu_improved_2024}. Among MLLMs, adapter style architectures are notable for their simplicity and have demonstrated state-of-the-art performance by combining a pretrained Vision Encoder with a pretrained Large Language Model (LLM) through a learned adapter network, that maps vision embeddings into the LLM’s token space \cite{merullo_linearly_2023, liu_visual_2023, liu_improved_2024}. These capabilities have motivated a growing body of work that adapts MLLMs for pixel-level semantic segmentation tasks, arguing that language supervision can improve interpretability and enable compositional segmentation driven by Referring Expression Segmentation, which allows segmentation through instruction following beyond what traditional segmentation models support \cite{hu_segmentation_2016, lai_lisa_2024}.

While recent works have proposed MLLM-based methods specialized for semantic segmentation \cite{xia_gsva_2024,chen_sam4mllm_2024,ren_pixellm_2024,zhang_omg-llava_2024}, it remains unclear whether the MLLM as a whole is actually better at image segmentation than its underlying vision encoder under a controlled probing setup. Recent analyses suggest that, without task-specific tuning, MLLMs may underperform on classical vision tasks and may overlook visual evidence that is already present in their vision backbones \cite{zhang_why_2024,fu_hidden_2025,tong_eyes_2024,fu_blink_2024}.

To address this gap in the literature, we propose a systematic, intervention-driven framework to dissect where segmentation competence arises or degrades across the MLLM stack. First, we perform linear probing of frozen embeddings at the vision encoder, the adapter, and every intermediate LLM layer (\cref{sec:probing}). This reveals a \emph{representation drop-off} at the adapter: projecting into the LLM embedding space trades fine-grained spatial fidelity for cross-modal alignment, degrading token-level separability. However, downstream LLM layers progressively \emph{self-refine} these visual representations and gradually recover segmentation quality.

This recovery raises the question of whether it is actively driven by cross-token attention or is merely a byproduct of residual connections and normalization. To further study these mechanisms, we design attention knockout interventions on conflicting class pairs (e.g., ceiling vs.\ sky), which suppress specific attention pathways while leaving all other operations unchanged (\cref{sec:knockout}). We find that correctly classified tokens act as semantic anchors whose attention signals pull misclassified neighbors toward the correct label, indicating that cross-token attention drives the observed self-refinement.

Finally, under standard causal attention, early image tokens can only attend to preceding tokens in sequence order and thus lack global spatial context, limiting self-refinement at those positions. We ablate this context starvation by introducing bidirectional attention exclusively among image tokens and by varying the vision encoder (\cref{sec:bidi}). Bidirectional attention eliminates the positional bias at early patches, and the degree of recovery depends on the compatibility between the vision encoder’s representation space and the LLM’s embedding space, with text-aligned encoders (e.g., CLIP, SigLIP) benefiting more than vision-only ones (e.g., DINOv2).

Our contribution can be summarized as follows (\cref{fig:main}):
\begin{itemize}
    \item We reveal a representation drop-off at the adapter and progressive self-refinement across LLM layers by performing layerwise linear probing that analyzes segmentation competence across the MLLM stack. 
    \item Our attention knockout experiments show that self‑refinement is driven by cross‑token attention, with correctly classified tokens acting as semantic anchors that guide their misclassified neighbors.
    \item We further find that applying bidirectional attention alleviates context starvation at early image tokens and that text-aligned encoders benefit more from LLM-layer refinement than vision-only encoders.
\end{itemize}

Across experiments, we observe that LLM layers provide structure- and constraint-aware refinement that improves local consistency and resolves certain class conflicts, but they do not recover fine-grained spatial details absent from the encoder. These findings clarify when and where MLLMs aid semantic segmentation and where they fall short, offering practical guidance for the design of segmentation-capable MLLMs.
\section{Related Work}\label{sec:related}
\textbf{Vision Encoders for Segmentation.} Vision Transformers pretrained at scale with supervised, contrastive, or self-supervised objectives have become powerful feature extractors for dense prediction tasks \cite{dosovitskiy_image_2021, radford_learning_2021,ravi_sam_2024,caron_emerging_2021, zhai_sigmoid_2023}. Notably, patch-level features from these models already transfer strongly to segmentation and other dense tasks when decoded with lightweight heads or linear probes \cite{banani_probing_2024,oquab_dinov2_2024}, establishing that rich spatial information is present in the encoder representations themselves. Our work builds on this probing paradigm but extends it beyond the vision encoder in isolation: we probe representations at every stage of the full MLLM pipeline, from the encoder through the adapter and into each LLM layer, to understand whether the downstream components preserve, degrade, or enhance the spatial information already present in the encoder features.

\textbf{MLLMs for Segmentation.} Adapter-style MLLMs connect a pretrained vision encoder to a pretrained LLM through a learned projection \cite{merullo_linearly_2023, liu_visual_2023,yao_dense_2024}, enabling multimodal reasoning without training either component from scratch. A growing line of work adapts this architecture for segmentation, leveraging instruction following and language-conditioned control to produce pixel-level outputs. A common strategy introduces special tokens that bridge language reasoning with mask prediction. LISA \cite{lai_lisa_2024} pioneered this by connecting an MLLM to SAM via a learned \texttt{[SEG]} token, enabling reasoning segmentation from complex language instructions. GSVA \cite{xia_gsva_2024} extended this to multi-target and empty-target cases with multiple \texttt{[SEG]} and \texttt{[REJ]} tokens. PixelLM \cite{ren_pixellm_2024} takes a SAM-free approach, generating multi-scale segment tokens decoded by a lightweight head. Other recent notable efforts include GLaMM \cite{rasheed_glamm_2024}, SAM4MLLM \cite{chen_sam4mllm_2024}, OMG-LLaVA \cite{zhang_omg-llava_2024}, and others \cite{zhang_psalm_2024, zou_segment_2023, tong_cambrian-1_2024,zhang_groundhog_2024,wu_f-lmm_2025}, each proposing architectural variants for grounding or segmentation.
While these works report competitive segmentation results, they focus on system-level performance and do not investigate where within the MLLM stack segmentation competence actually resides. Our work complements this line of research by providing a diagnostic analysis of the representations these architectures produce.

\textbf{Diagnostic Analyses of MLLM Representations.}
Recent analyses caution that MLLMs may underperform on standard vision tasks without task-specific finetuning. Zhang \etal~\cite{zhang_why_2024} examine the performance of MLLMs on classification tasks and find that they can overlook visual evidence encoded by their own vision backbone. Other works in this area have shown that MLLMs substantially underperform their vision encoders on vision-centric tasks such as depth estimation and correspondence, identifying the LLM as the primary bottleneck. Their analysis probes representations across the VLM but focuses on general perceptual tasks and does not perform causal interventions \cite{fu_hidden_2025}. Conversely, Li \etal~\cite{li_exploring_2025} show that generative MLLMs can extract visual information more effectively than CLIP from the same frozen encoder, suggesting a more nuanced picture. Liang \etal~\cite{liang_unleashing_2024} demonstrate that intermediate MLLM layers can hold richer region-level descriptions than the final layer. We build on these findings but focus specifically on semantic segmentation, introduce attention knockout experiments to test whether LLM layers actively resolve ambiguity, and ablate architectural choices such as attention directionality and vision encoder selection. Our work contributes a segmentation-focused, causal-intervention view, complementing classification- and correspondence-oriented findings.

\section{Layerwise Linear Probing}
\label{sec:probing}

\begin{figure*}[tb]
    \centering
    \includegraphics[width=\textwidth]{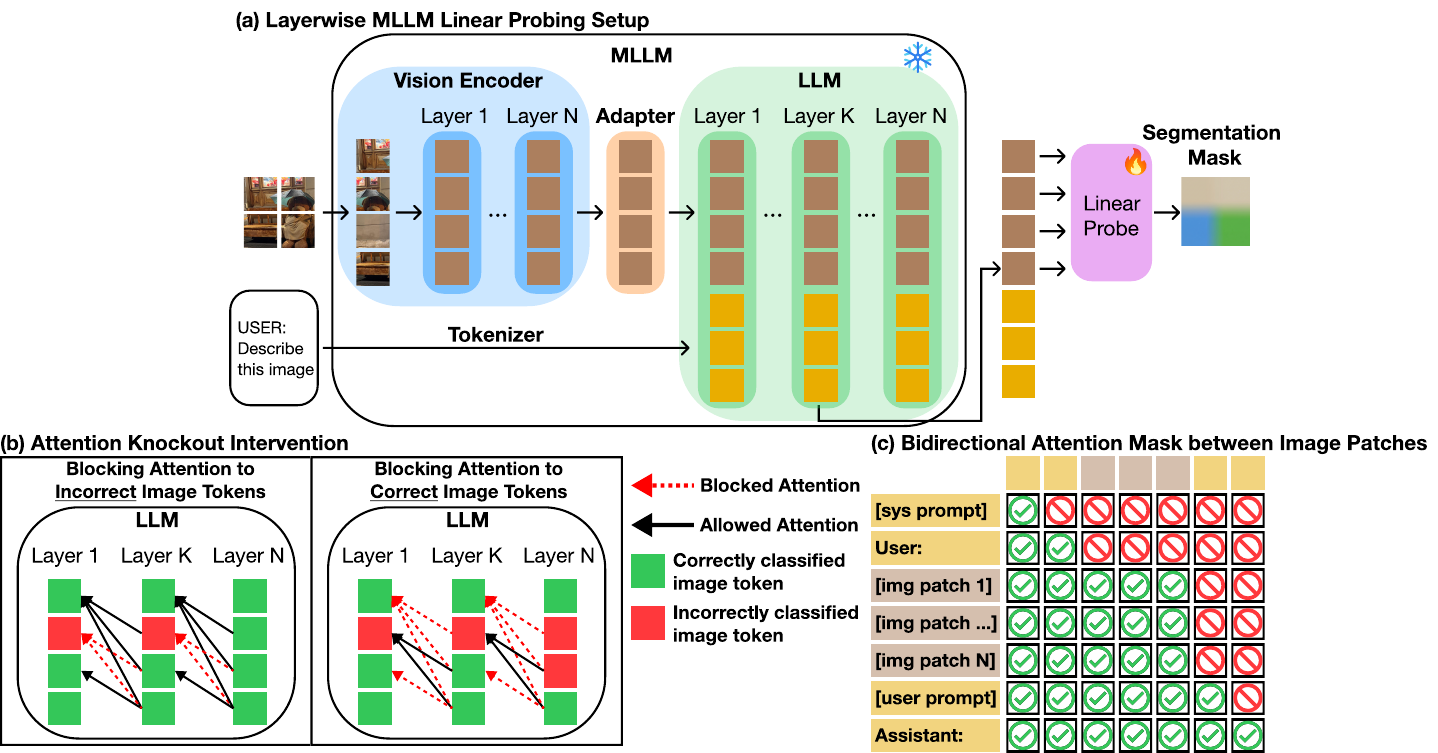}
    \caption{\textbf{Overview of the three analysis methods.} \textbf{(a)}~Layerwise linear probing: given an input image, the vision encoder produces patch token embeddings (brown), which are projected by the adapter into the LLM's embedding space. Inside the LLM, image tokens are processed jointly with text prompt tokens (yellow). At each layer $\ell$, we extract only the image token representations and train an independent linear probe to predict per-patch semantic classes, reassembled into a 2D segmentation map. \textbf{(b)}~Attention knockout: we selectively block attention to incorrectly classified tokens (left) or correctly classified tokens (right) across all LLM layers, testing whether cross-token attention drives self-refinement. \textbf{(c)}~Bidirectional attention mask: image tokens attend to each other bidirectionally while all other token pairs retain causal masking, alleviating context starvation at early image positions.}
    \label{fig:method_overview}
\end{figure*}

\textbf{Models and notation.}
Given an input image, the vision encoder divides it into a grid of $T$ non-overlapping patches (e.g., $T=576$ for a $336\times336$ image with patch size 14) and maps them to a sequence of token embeddings. The adapter, a learned two-layer MLP, projects each token embedding into the LLM's $d$-dimensional input space. Within each LLM layer, the full input sequence contains system tokens, image tokens, and text prompt tokens. We denote by $\mathbf{X}^{(\ell)} \in \mathbb{R}^{T \times d}$ the representations corresponding to the $T$ image patch tokens only at layer $\ell$. Under standard causal attention, image tokens can only attend to preceding tokens in the sequence (i.e., system tokens and earlier image patches) and are therefore not influenced by the text prompt that follows them.
Since each token corresponds to a fixed spatial position in the image, per-token predictions can be reshaped into a 2D grid and upsampled to the original resolution for pixel-level evaluation.

\textbf{Datasets.}
We evaluate on the following three standard segmentation benchmarks: ADE20K (150 classes, diverse indoor and outdoor scenes) \cite{zhou_semantic_2018}; PASCAL VOC 2012 (20 foreground classes, augmented training set) \cite{everingham_pascal_2010}; and Cityscapes (urban street scenes) \cite{cordts_cityscapes_2016}.
We use standard train/validation splits and report mIoU as the primary metric and pixel accuracy (pAcc) as a secondary measure.

\textbf{Probing protocol.}
We introduce a probing protocol to systematically evaluate where segmentation competence arises, degrades, or is refined across the MLLM stack. Our framework targets adapter-style MLLMs, which combine a frozen vision encoder, a trained adapter, and an LLM. This modular structure allows us to isolate and compare representations at three stages: the vision encoder output, the adapter output, and each intermediate LLM layer.
To evaluate the segmentation quality of representations at each stage of the MLLM, we train an independent linear probe per layer \cite{alain_understanding_2018}. The procedure is identical across all stages for the vision encoder, adapter, and LLM layers, and differs only in which hidden state is extracted and the dimension of the hidden state.

For a target layer $\ell$, we freeze the entire MLLM and extract $\mathbf{X}^{(\ell)}$ for every image in the training set. Each token is independently classified by a linear probe trained with cross-entropy on the frozen features. Additional training and implementation details are provided in the supplementary material.

We train three MLLM variants by pairing Vicuna-7B \cite{touvron_llama_2023} with CLIP ViT-L/14@336, DINOv2 Large@336, and SigLIP SO400M/14, each following the standard LLaVA-1.5 two-stage procedure \cite{liu_improved_2024}: pretraining the adapter on 558K image-caption pairs, then finetuning the full model on 665K visual instruction data.
By comparing mIoU across layers under identical probe training conditions, we obtain a complete profile of how segmentation-relevant information evolves from the vision encoder through the adapter and into the LLM.

\subsection{The Adapter Introduces a Representation Drop-off}
\label{subsec:dropoff-recovery}

We first compare the segmentation quality of features immediately before and after the adapter. \cref{fig:layerwise-recovery} reports the mIoU for three vision encoders at the encoder output and the adapter output. Across all encoders, we observe a consistent drop in mIoU after the adapter projects the visual features into the LLM's embedding space. The magnitude of this drop varies: CLIP experiences a modest decline, DINOv2 and SigLIP suffer larger degradations. This representation drop-off indicates that the adapter introduces a structural bottleneck, trading fine-grained spatial fidelity for cross-modal alignment with the language embedding space. Analogous representation gaps have been documented in contrastive vision-language spaces, where image and text embeddings occupy geometrically separated regions \cite{liang_mind_2022}, while Fu~\etal~\cite{fu_hidden_2025} show that VLMs can underperform their own vision encoders on tasks such as correspondence.

\subsection{LLM Layers Progressively Recover Segmentation Quality}

\begin{figure}[tb]
    \centering
    \includegraphics[width=\linewidth]{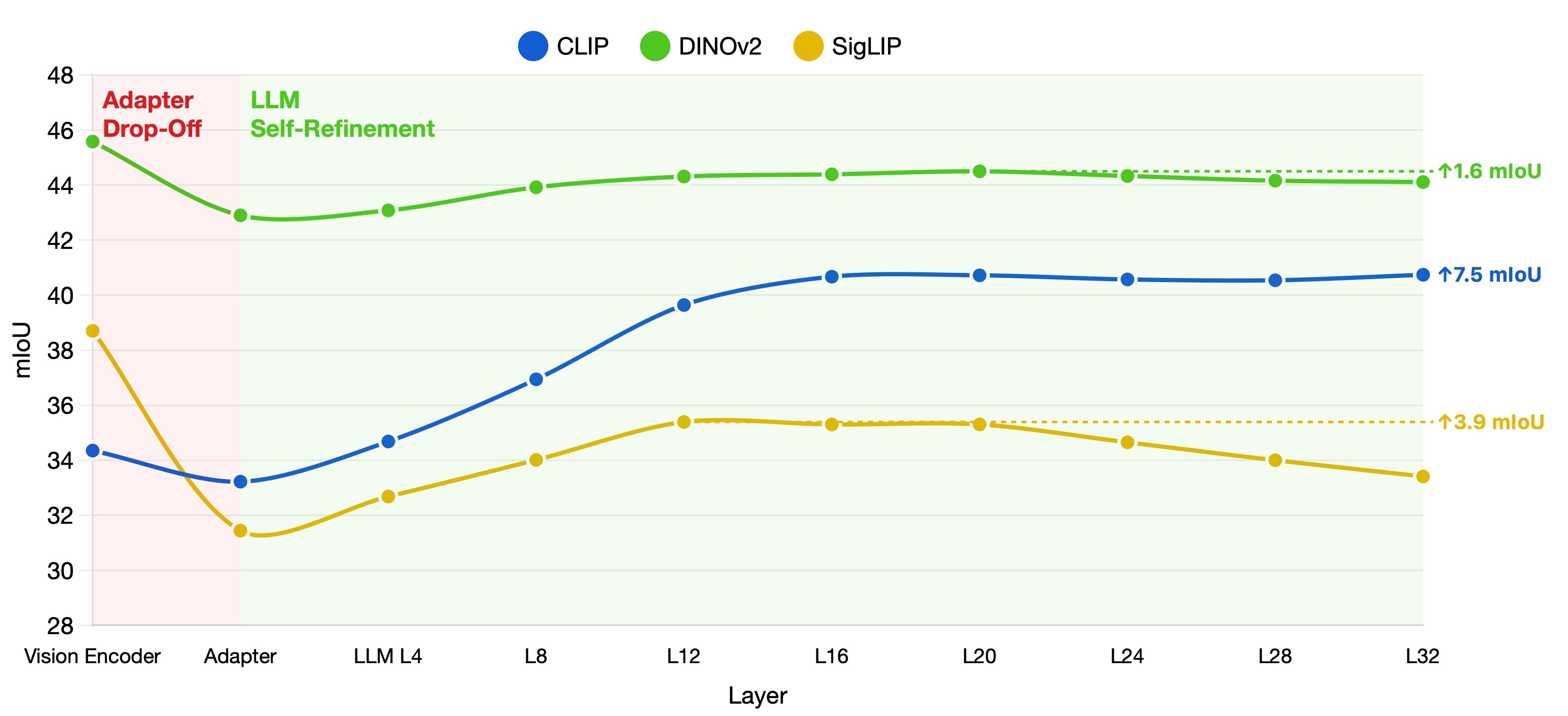}
    \caption{\textbf{Layerwise linear probing results across the MLLM stack.} mIoU on ADE20K for CLIP, DINOv2, and SigLIP encoders paired with Vicuna-7B, measured at the vision encoder output, adapter output, and each LLM layer. All three encoders exhibit a drop at the adapter followed by progressive recovery across LLM layers. Dashed lines mark the best-performing layer; values on the right indicate the total mIoU improvement across LLM layers.}
    \label{fig:layerwise-recovery}
\end{figure}

Segmentation quality does not continue to degrade as features propagate through the LLM. On the contrary, as shown in \cref{fig:layerwise-recovery}, mIoU steadily recovers across the LLM layers. The recovery is characterized by a sharp increase in mIoU in the early LLM layers, followed by a plateau in the mid-to-late layers where peak performance is reached. This is a notable finding: The LLM layers, which were pretrained for language modeling and not for spatial reasoning, are able to refine the visual representations and restore segmentation-relevant structure that was lost at the adapter.

The drop-off and recovery pattern is consistent across all three encoders, but the strength of recovery varies. CLIP, which is pretrained with a contrastive objective that aligns visual features to text, exhibits the strongest recovery: its LLM-layer representations ultimately exceed the vision encoder baseline. SigLIP, which shares a similar contrastive pretraining objective, also shows meaningful recovery. DINOv2, a vision-only encoder with no text alignment, recovers less strongly. This suggests that the degree of compatibility between the vision encoder's representation space and the LLM's embedding space influences how effectively the LLM layers can refine the visual features. 

\subsection{Qualitative Evidence and Semantic Clustering}

\begin{figure}[tb]
    \centering
    \includegraphics[width=\linewidth]{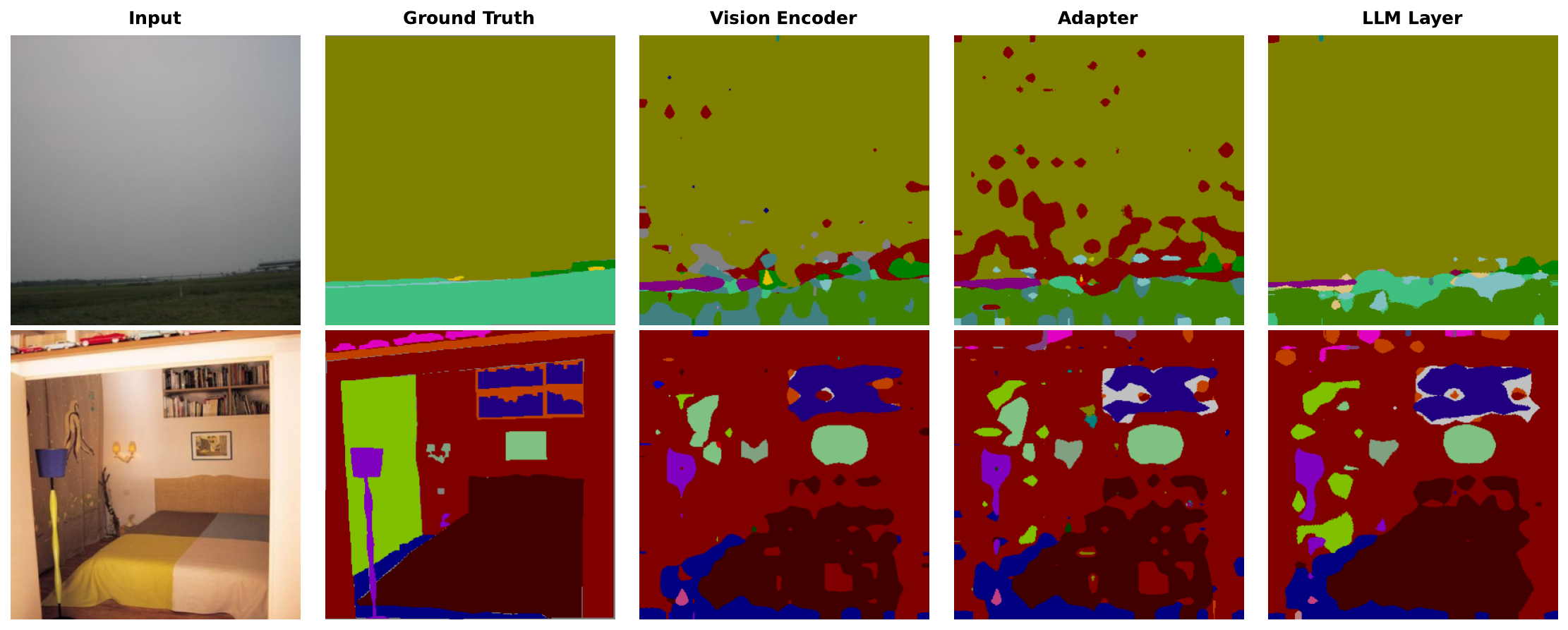}
    \caption{\textbf{Qualitative segmentation predictions across the MLLM stack.} From left to right: input image, ground truth, linear probe prediction at the vision encoder output, adapter output, and at an intermediate LLM layer. Representation drop-off at the adapter but deeper layers appear to resolve class confusions (e.g., wall vs.\ bed) and produce more spatially coherent predictions.}
    \label{fig:qualitative-probing}
\end{figure}

\cref{fig:qualitative-probing} shows segmentation predictions from the linear probe at different stages of the MLLM for representative ADE20K validation images. At the adapter output, predictions exhibit noisy boundaries and more frequent confusion between classes. At deeper LLM layers, these errors are progressively resolved: boundaries become more spatially coherent and class assignments stabilize. This visual evidence complements the quantitative mIoU improvements and suggests that the LLM layers perform a form of contextual refinement, leveraging global token interactions and semantics to re-impose structure to disambiguate local patch-level predictions, yielding net gains for specific conflicts.

\begin{figure}[tb]
    \centering
    \includegraphics[width=\linewidth]{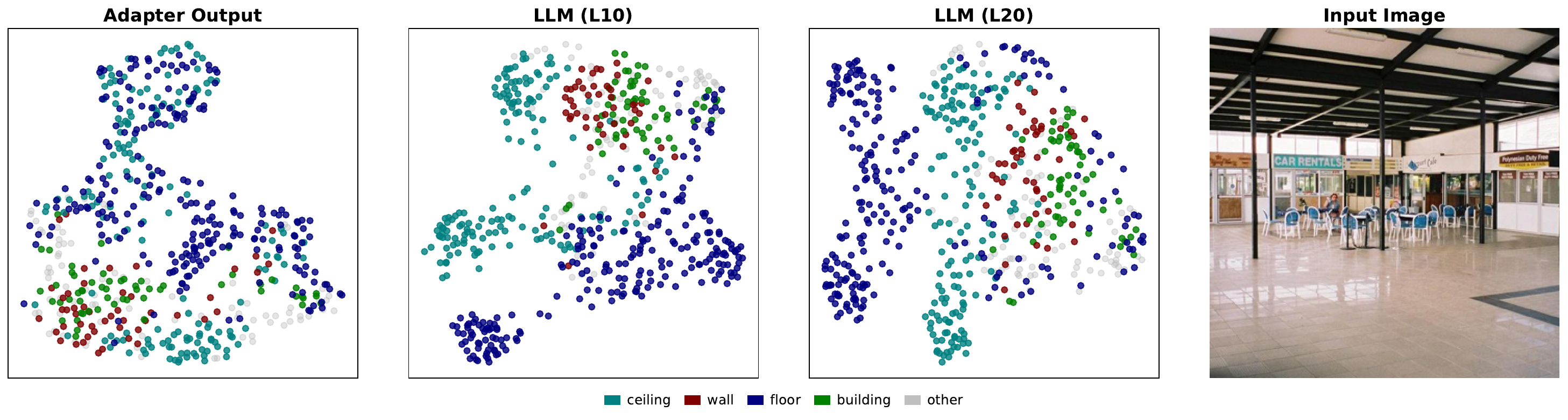}
    \caption{\textbf{UMAP projections of patch-level hidden states across the MLLM stack.} Each point represents one of 576 image patch tokens, colored by semantic class. At the adapter output, classes are interleaved, by layer 20, same-class patches form distinct clusters, illustrating the progressive emergence of semantic structure through the LLM layers.}
    \label{fig:umap-layers}
\end{figure}

To further investigate how the LLM layers refine visual representations, we visualize the hidden states of individual image patches using UMAP projections at different depths of the model.
\cref{fig:umap-layers} shows the 2D embeddings of all 576 patch tokens from a single image of the CLIP MLLM at the adapter output, an intermediate LLM layer, and the layer at which linear probing performance peaks.
Each patch is colored by the semantic class and classes not among the four most prevalent are shown in gray.
At the adapter output, patches from different semantic classes are heavily interleaved, confirming that the projected features lack clear category-level organization.
As representations pass through the LLM, same-class patches progressively cluster together, and by layer 20, distinct semantic groups such as floor, ceiling, wall, and building occupy clearly separated regions of the embedding space, wall and building cluster close together because of semantic similarity.
This provides a complementary, geometric perspective on the recovery and corroborates the mIoU improvements observed in the linear probing results and qualitative results: The LLM does not merely make features more linearly separable, but actively re-organizes them into semantically coherent clusters.

\section{Attention Knockout}
\label{sec:knockout}

The layerwise probing results from \cref{subsec:dropoff-recovery} show that segmentation quality improves across LLM layers, suggesting that the model progressively resolves classification errors. To test whether this self-refinement is actively mediated by cross-token attention rather than being a passive byproduct of residual connections or layer normalization, we design a global attention knockout experiment. We adapt the attention knockout technique introduced by Geva \etal~\cite{geva_dissecting_2023} for tracing information flow in auto-regressive language models, and applied to MLLM by Neo~\etal~\cite{neo_towards_2025} to study how visual information is extracted at the output position. While both prior works use knockout to identify which tokens inform the model's generated text output, we repurpose the technique to probe a different question: whether cross-token attention among image patch tokens themselves drives the self-refinement of spatial representations across LLM layers.

\textbf{Procedure.}
Given a test image, we first obtain predicted labels $\hat{y}_t$ for each image patch token $t$. We select a target class $c$ to block and identify the corresponding token set $\mathcal{B}_c = \{s \in \mathcal{I} : \hat{y}_s = c\}$, where $\mathcal{I}$ denotes the full set of image tokens. At every LLM layer $\ell$, we mask out all attention from any image token to any token in $\mathcal{B}_c$ by setting the pre-softmax attention logits to $-\infty$:
\begin{equation}
\mathbf{A}^{(\ell)}_{t \leftarrow s} \leftarrow -\infty \quad \forall\; t \in \mathcal{I},\; s \in \mathcal{B}_c.
\label{eq:knockout}
\end{equation}
This renders class $c$ completely invisible: No image token, including tokens of class $c$ themselves, can attend to the blocked tokens. Blocked tokens can still attend to all non-blocked tokens, so their representations continue to evolve but without any self-reinforcement from same-class neighbors. We then extract hidden states at every layer and evaluate segmentation via the same per-layer linear probes used in the earlier experiment.

\textbf{Experimental conditions.}
We focus on images where the unmodified model exhibits characteristic class confusions, e.g., patches of ceiling misclassified as sky. For each such image, we run two complementary conditions and compare the resulting layerwise segmentation against the unmodified model (\cref{fig:method_overview}b):
\begin{enumerate}
    \item \textbf{Block incorrect class.} We block the class that the model \emph{incorrectly} assigns to some tokens (e.g. block \emph{sky} when the ground truth is \emph{ceiling}). If attention to incorrectly classified tokens reinforces errors, removing their influence should accelerate self-correction through global context.
    \item \textbf{Block correct class.} We block the class that is \emph{correctly} assigned (e.g. block \emph{ceiling}). If correctly classified tokens serve as semantic anchors in a global context that pull misclassified neighbors toward the right label, removing them should impair self-correction.
\end{enumerate}

\subsection{Blocking the Incorrect Class Accelerates Self-Correction}
\label{subsec:knockout-results}

We select ADE20K validation images where the unmodified model exhibits characteristic class confusions, such as ceiling patches misclassified as sky, and compare layerwise segmentation maps under the two blocking conditions against the unmodified model.

\begin{figure}[tb]
    \centering
    \includegraphics[width=\textwidth]{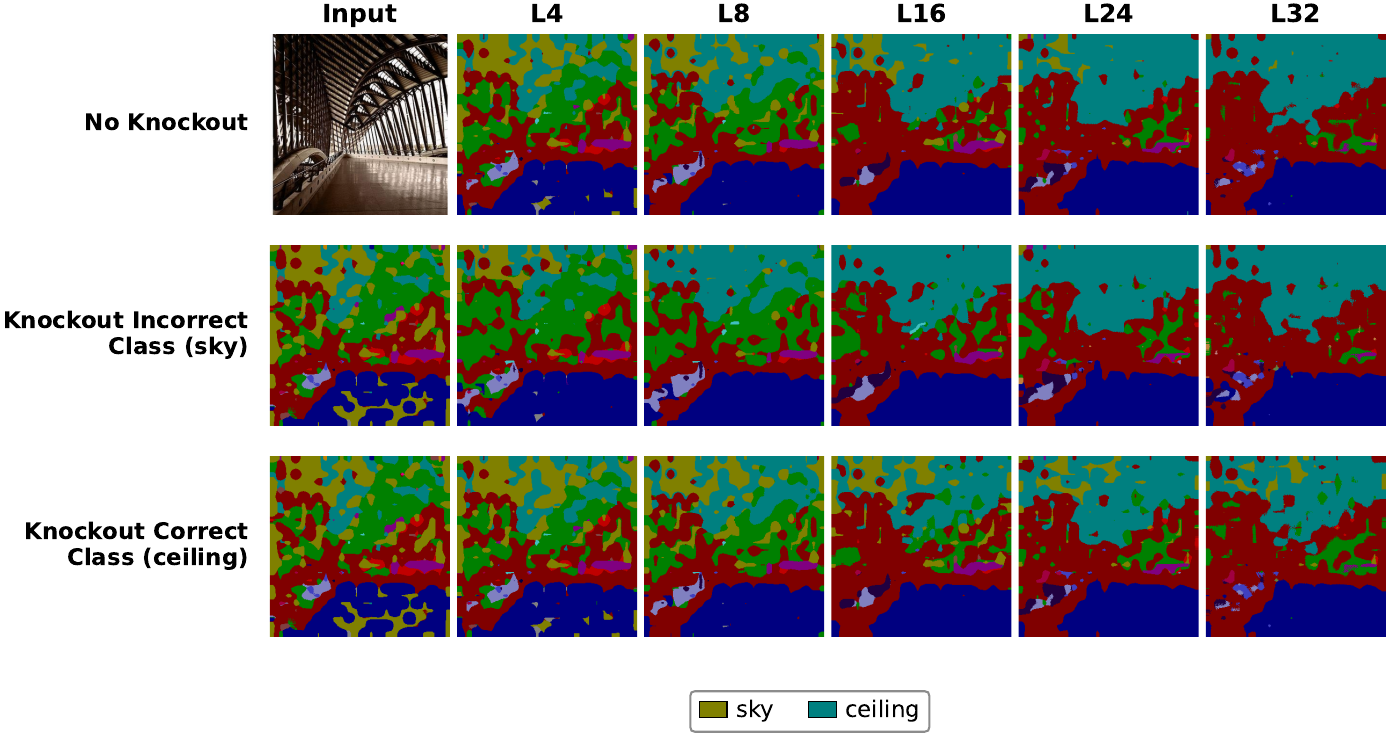}
    \caption{\textbf{Global attention knockout: layerwise segmentation comparison.} Top row: unmodified baseline. Middle row: incorrect class blocked (e.g. sky). Bottom row: correct class blocked (e.g. ceiling). Blocking the incorrect class accelerates the resolution of misclassified patches across layers, while blocking the correct class impairs self-correction and leaves more errors in mid-to-late layers.}
    \label{fig:knockout-comparison}
\end{figure}

When we block the incorrectly assigned class (e.g. making \emph{sky} invisible in an image where the ground truth is \emph{ceiling}), misclassified tokens are corrected faster across layers compared to the unmodified model. Already in the early LLM layers, the segmentation maps show reduced confusion in the affected region, and by the final layer the model, with incorrect classes knocked out, produces fewer residual misclassifications than the unmodified model. This indicates that tokens carrying the incorrect label were actively reinforcing erroneous predictions through attention: by silencing them, the remaining contextual cues from correctly classified neighbors (e.g. walls, floors, and other indoor elements) dominate the attention field, enabling the model to resolve the ambiguity more efficiently.

\subsection{Blocking the Correct Class Impairs Self-Correction}

The complementary condition reveals the opposite effect. When we block the correctly assigned class (e.g. making ceiling invisible), the model's ability to self-correct deteriorates markedly. Misclassified tokens persist longer across the layer progression, and in the mid-to-late layers the segmentation quality falls below that of the unmodified model, with more misclassified patches remaining than when no intervention is applied. This demonstrates that correctly classified tokens serve as semantic anchors: Their attention signals help pull misclassified neighbors toward the correct label. Without these anchors, the model loses its primary self-correction mechanism and errors are left unresolved or even amplified.

\subsection{Cross-Token Attention Drives Self-Refinement}

Together, these two conditions provide direct evidence that the representation refinement documented in \cref{subsec:dropoff-recovery} is not a passive byproduct of residual connections or layer normalization, but is actively mediated by cross-token attention. The LLM layers leverage semantic context, attending to tokens of the correct class, to progressively resolve local classification errors. However, because self-correction relies on the presence of correctly classified anchors, this mechanism cannot recover fine-grained spatial details absent from the encoder features or overcome systematic encoder biases where no correct anchors exist. These findings confirm the self-refinement hypothesis and identify cross-token attention as its operative mechanism.
\section{Causal vs.\ Bidirectional Attention}
\label{sec:bidi}

The preceding experiments established that cross-token attention drives self-refinement across LLM layers (\cref{subsec:dropoff-recovery,subsec:knockout-results}).
However, under standard causal attention, image tokens are processed in raster order (left-to-right, top-to-bottom) and each token can only attend to preceding tokens in the sequence.
The first image token cannot see any other image tokens, while the last token attends to all $T$ image patches.
This creates a structural asymmetry in which early tokens lack global spatial context, a limitation particularly relevant for segmentation, where every patch should ideally access scene-level information.
We observed this effect in our layerwise probing experiments: tokens in the first row, and especially the top-left corner, are often misclassified in the qualitative visualizations and exhibited consistently lower classification accuracy that did not improve across LLM layers.

\textbf{Image-only bidirectional attention.}
To test whether this positional bias limits segmentation quality, we modify the attention mask to grant bidirectional attention exclusively among image tokens while preserving causal attention for text generation (\cref{fig:method_overview}c).
Let $\mathcal{I}$ denote the set of image token positions within the input sequence.
The modified mask function is:
\begin{equation}
M(q, k) = \underbrace{(q \in \mathcal{I} \;\wedge\; k \in \mathcal{I})}_{\text{image--image: bidirectional}} \;\;\vee\;\; \underbrace{(q \geq k)}_{\text{causal}},
\label{eq:bidi-mask}
\end{equation}
where $q$ and $k$ index the query and key positions in the full input sequence, respectively.
When both tokens lie within the image region, attention is permitted regardless of their relative position, but for all other pairs, the standard causal constraint applies.
This design differs from the prefix-LM strategy employed by PaLiGemma~\cite{beyer_paligemma_2024}, which grants bidirectional attention across \emph{all} input tokens, image, system prompt, and task prefix alike.
Our variant targets spatial self-refinement among image tokens specifically, isolating its effect on segmentation while preserving the sequential structure required for autoregressive text generation.

\textbf{Training.}
We follow the standard LLaVA-1.5 two-stage procedure: (1) pretraining the adapter on 558K image-caption pairs with the LLM frozen, and (2) finetuning the full model on 665K visual instruction data.
Both stages use the same attention type, either fully causal or fully bidirectional, so that the comparison reflects the cumulative effect of attention directionality across the entire training process.
All MLLM hyperparameters, training data, and vision encoders (CLIP, DINOv2, SigLIP) are identical between the two conditions, only the attention mask differs.

\subsection{Early Tokens Suffer Context Starvation}
\label{subsec:bidi-results}

\begin{figure}[tb]
    \centering
    \includegraphics[width=\textwidth]{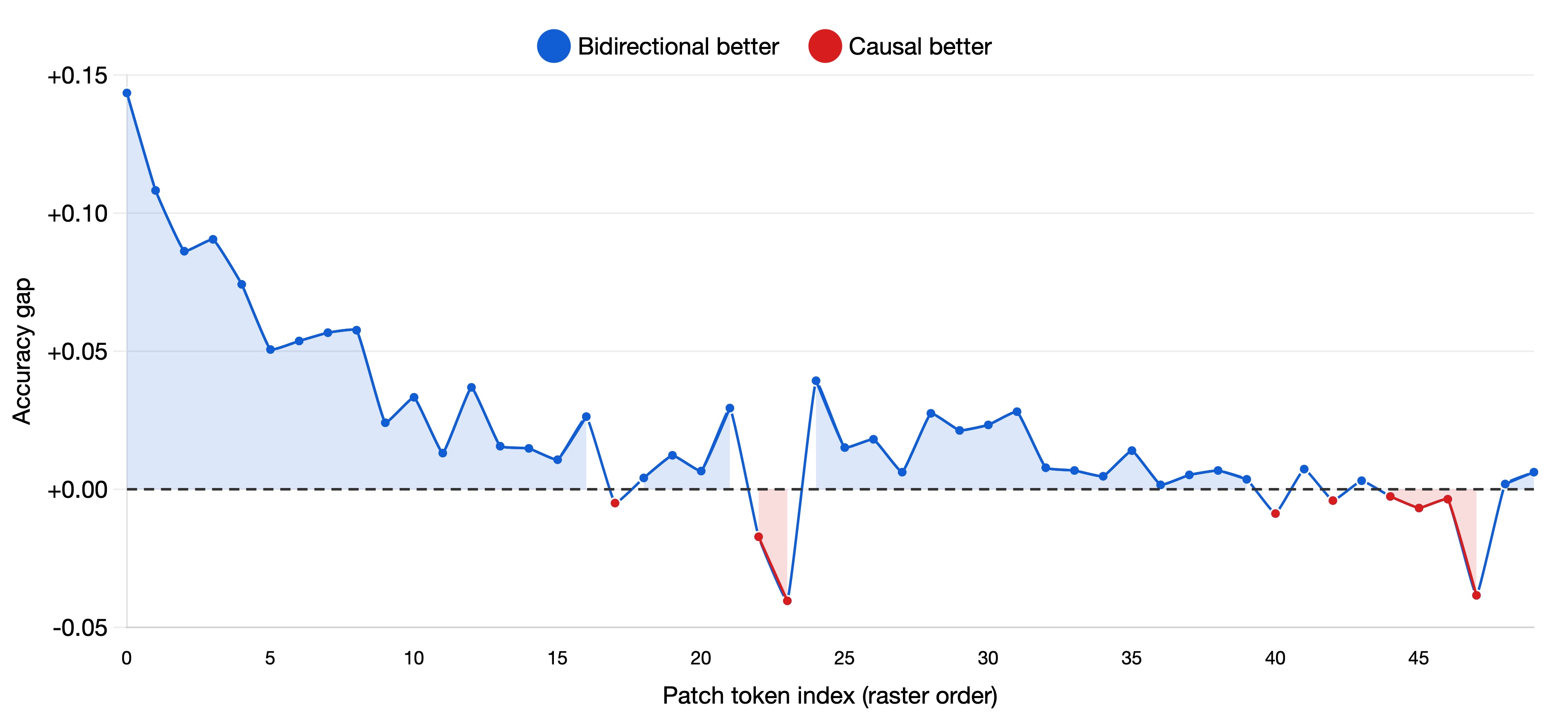}
    \caption{\textbf{Context starvation under causal attention.} Per-patch pixel accuracy for the first 50 tokens in the LLM layer of the MLLM. Accuracy Gap: Bidirectional minus causal accuracy. Early patches suffer severe context starvation under causal masking.}
    \label{fig:context_starvation_patch}
\end{figure}

We compare per-patch classification accuracy between the causal and bidirectional CLIP MLLM models at the same layer across the ADE20K validation set.
To isolate the positional effect of the attention mask from the content-distribution bias shared by both models, we report the difference in pixel accuracy between the bidirectional and causal attention mechanisms evaluated at patch positions.
\cref{fig:context_starvation_patch} reports accuracy for the first 50 patch tokens in patch order.
The gap is striking: Patch 0, which under causal attention attends to no visual context, shows a $+$14.35 percentage-point pixel accuracy increase, a 23.2\% relative improvement under bidirectional attention; the gap decays in later image patches.

\subsection{Bidirectional Attention Sustains Recovery Across Layers}

\begin{figure}[tb]
    \centering
    \includegraphics[width=\textwidth]{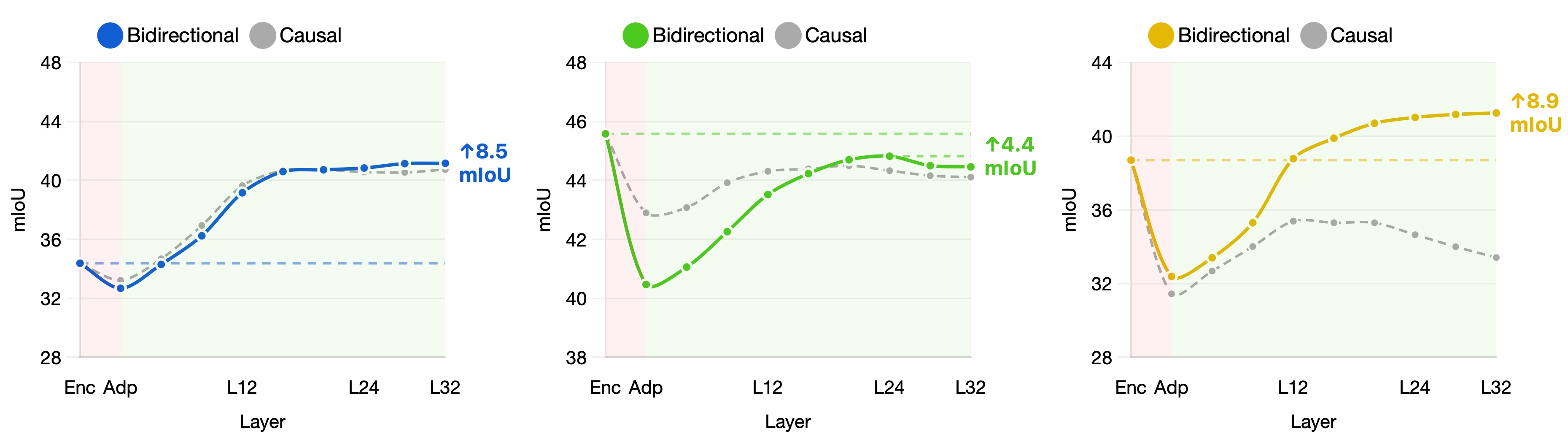}
    \caption{\textbf{Causal vs. bidirectional attention: Layerwise probing on ADE20K across the MLLM Stack for different vision encoders.} Bidirectional attention yields modest peak mIoU gains for CLIP and DINOv2, while SigLIP shows a larger effect.}
    \label{fig:bidi_vs_causal_triptych}
\end{figure}

\cref{fig:bidi_vs_causal_triptych} compares layerwise linear probing mIoU for causal and bidirectional MLLMs across all three encoder configurations.
Bidirectional attention yields modest peak mIoU gains for CLIP ($+$0.42) and DINOv2 ($+$0.32), consistent with the per-patch analysis: The improvement is concentrated at the few context-starved positions rather than reflecting a broad representational change.
SigLIP shows a larger effect: Under causal attention, recovery peaks at layer 12 (mIoU 35.39) and then declines through deeper layers, while bidirectional attention sustains monotonic improvement to 41.26 at layer 32.

\subsection{Self-Refinement Requires Access to Neighbors}

The bidirectional attention experiment reveals context starvation as a real but sharply localized cost of causal masking.
For the first few image tokens, the causal mask creates a persistent representation penalty that is not resolved across LLM layers; bidirectional attention eliminates it by granting immediate access to all visual neighbors.
Beyond these early positions, causal and bidirectional attention produce comparable representations, indicating that self-refinement through partial context is sufficient once a token has access to even a modest number of visual neighbors.
This finding reinforces the self-refinement narrative from \cref{subsec:knockout-results}: Self-refinement depends on access to semantically consistent neighbors, and causal masking delays the formation of these contextual anchors for early tokens, whereas bidirectional attention restores immediate global context and enables refinement to operate uniformly across spatial positions.

\subsection{Effect on Language Understanding}

The preceding sections show that bidirectional attention among image tokens improves segmentation, particularly for context-starved early patches. A natural concern is whether this modification degrades the model's language capabilities. \cref{tab:bidi-vqa} compares VQA performance across nine benchmarks for causal and bidirectional variants of all three encoder configurations.
CLIP and SigLIP models maintain comparable performance to their causal baselines across most benchmarks, indicating that bidirectional image attention preserves language understanding. DINOv2 shows minor regressions, consistent with its weaker text alignment observed throughout our experiments. These results suggest that bidirectional attention among image tokens offers a favorable trade-off: it alleviates context starvation for segmentation without sacrificing language capabilities for text-aligned encoders.

\begin{table*}[tb]
\centering
\caption{\textbf{VQA benchmark: causal vs.\ bidirectional LLaVA\,1.5 (7B) across three vision encoders.} All models use Vicuna-7B as the LLM backbone. Bidirectional attention preserves language understanding with segmentation gains that vary across encoders. DINOv2 shows broader regressions, consistent with its lack of text alignment.}
\label{tab:bidi-vqa}
\resizebox{\textwidth}{!}{%
\begin{tabular}{@{}ll cccc cccc c@{}}
\toprule
Vision Encoder & Attention & GQA & MMB & MME\textsuperscript{P} & MME\textsuperscript{C} & MMMU & POPE & SQA\textsuperscript{I} & TextVQA & VizWiz \\
\midrule
\multirow{2}{*}{CLIP ViT-L/14}
 & Causal        & 62.6 & 66.4 & 1483 & 284 & 35.3 & 86.8 & 68.7 & 46.9 & 56.0 \\
 & Bidirectional & 62.7 & 65.5 & 1538 & 288 & 36.2 & 86.9 & 69.7 & 47.0 & 57.5 \\
\midrule
\multirow{2}{*}{DINOv2 ViT-L/14}
 & Causal        & 62.1 & 57.7 & 1304 & 324 & 34.6 & 87.2 & 66.1 & 14.0 & 51.4 \\
 & Bidirectional & 60.5 & 55.3 & 1247 & 326 & 32.4 & 85.1 & 66.3 & 13.7 & 45.8 \\
\midrule
\multirow{2}{*}{SigLIP SO400M/14}
 & Causal        & 61.1 & 63.7 & 1414 & 275 & 34.7 & 84.4 & 70.4 & 50.2 & 58.2 \\
 & Bidirectional & 62.1 & 66.9 & 1402 & 298 & 34.9 & 84.8 & 70.7 & 53.6 & 53.3 \\
\bottomrule
\end{tabular}%
}
\end{table*}
\section{Conclusion}
\label{sec:conclusion}
Our probing and intervention framework reveals that adapter-style MLLMs introduce a representation drop-off at the adapter that degrades token-level separability, but LLM layers progressively recover segmentation quality through attention-mediated self-refinement. Attention knockout experiments confirm that correctly classified tokens act as semantic anchors whose attention signals pull misclassified neighbors toward the correct label, identifying cross-token attention as the operative refinement mechanism. Causal attention creates context starvation at early image token positions that bidirectional attention among image tokens alleviates by restoring immediate global context and enabling refinement to operate uniformly across spatial positions. These findings provide a mechanistic account of how MLLMs process visual information for segmentation, informing the design of future segmentation-capable MLLMs and hopefully providing a step for more interpretable multimodal systems.

\textbf{Limitations.}
We used LLaVA-type models as a starting point and varied the vision encoder across the configurations for interpreting MLLMs, but our conclusions might not generalize for significantly different architectures. Our linear probes may underestimate actual segmentation capacity achievable with richer task heads and the knockout interventions block attention globally across all layers, which does not isolate the contribution of individual layers to the refinement process. Future work could explore whether these findings extend to broader task types and model architectures.

\newpage
\section*{Acknowledgements}
Our work was partially funded by the ERC (853489 - DEXIM) and the Alfried Krupp von Bohlen und Halbach Foundation, which we thank for their support. 
The authors gratefully acknowledge the scientific support and resources of the AI service infrastructure LRZ AI Systems provided by the Leibniz Supercomputing Centre (LRZ) of the Bavarian Academy of Sciences and Humanities (BAdW), funded by Bayerisches Staatsministerium für Wissenschaft und Kunst (StMWK).

%
%
\bibliographystyle{splncs04}
\bibliography{main}

\newpage
\appendix

\section*{Supplementary Materials}

\section{Implementation Details}
\label{sec:supp-implementation}

\subsection{MLLM Training}
\label{subsec:supp-mllm-training}

All MLLM variants follow the LLaVA-1.5 two-stage training procedure~\cite{liu_improved_2024} using Vicuna-7B~\cite{touvron_llama_2023} as the LLM backbone.
We train six configurations: three vision encoders (CLIP ViT-L/14@336~\cite{radford_learning_2021}, DINOv2 ViT-L/14@336~\cite{oquab_dinov2_2024}, SigLIP SO400M/14@384~\cite{zhai_sigmoid_2023}) $\times$ two attention types (causal, bidirectional).
The adapter is a two-layer MLP that projects vision encoder outputs into the LLM's embedding space.
CLIP and DINOv2 operate at $336\times336$ resolution, producing $24\times24=576$ patch tokens; SigLIP operates at $384\times384$, producing $27\times27=729$ patch tokens.
All models are trained in bfloat16 precision.

\textbf{Stage~1 (Adapter pretraining).}
The adapter is trained on 558K image-caption pairs with both the vision encoder and LLM frozen.
We use AdamW with a learning rate of $1\times10^{-3}$, cosine learning rate schedule, warmup ratio of 3\%, and no weight decay and training runs for 1 epoch.

\textbf{Stage~2 (Visual instruction tuning).}
The full model (adapter + LLM) is finetuned on 665K visual instruction data, with the vision encoder remaining frozen.
We use AdamW with a learning rate of $2\times10^{-5}$, cosine schedule, warmup ratio of 3\%, and no weight decay and training runs for 1 epoch.

For the bidirectional attention variants (\cref{sec:bidi}), both stages use the bidirectional image-token attention mask defined in \cref{eq:bidi-mask}; the causal variants use the standard causal mask throughout.
All other hyperparameters are identical between causal and bidirectional conditions.

\subsection{Linear Probing Protocol}
\label{subsec:supp-probing}

We train an independent linear probe at each extraction point: the vision encoder output, the adapter output, and intermediate LLM layers.
The probe is a single linear layer mapping from the hidden dimension $d$ to $K$ classes, trained with plain cross-entropy loss without class reweighting or data augmentation.
We use the AdamW optimizer ($\beta_1=0.9$, $\beta_2=0.999$) with a learning rate of $1\times10^{-3}$, polynomial learning rate decay (power $0.9$), no weight decay, and a batch size of 64.
Each probe is trained for 20 epochs; we track validation mIoU after each epoch and retain the best-performing checkpoint.

The per-patch logits are reshaped into a 2D grid matching the original patch layout (e.g., $24 \times 24 \times K$ for 576 patches) and bilinearly upsampled to the full image resolution.
The loss is computed against the full-resolution ground-truth segmentation mask, allowing it to reflect sub-patch boundary information through the interpolated logits.
Images are processed in a single forward pass without sliding window or multi-scale evaluation.

The linear probe is trained exclusively on the training split and all reported metrics are computed on the held-out validation split, ensuring that the results reflect the generalization quality of the frozen representations rather than memorization by the probe.
We keep these training conditions across all layers and all experiments.

\subsection{Attention Knockout Setup}
\label{subsec:supp-knockout}

For the attention knockout experiments (\cref{sec:knockout}), we use the per-layer linear probes trained in the standard probing experiment.
We then identify the token set $\mathcal{B}_c$ corresponding to the class to be blocked and re-run the forward pass with the attention mask modified according to \cref{eq:knockout}, extracting hidden states at every layer for evaluation with the same frozen probes.
The knockout is implemented by registering forward pre-hooks on each LLM layer's self-attention module that set the corresponding pre-softmax attention logits to $-\infty$ before the softmax computation.
This ensures that any change in segmentation quality is attributable to the attention intervention.

\section{Extended Probing Results}
\label{sec:supp-probing-results}

\subsection{Layerwise Probing on PASCAL VOC and Cityscapes}

\begin{figure}[tb]
    \centering
    \includegraphics[width=\linewidth]{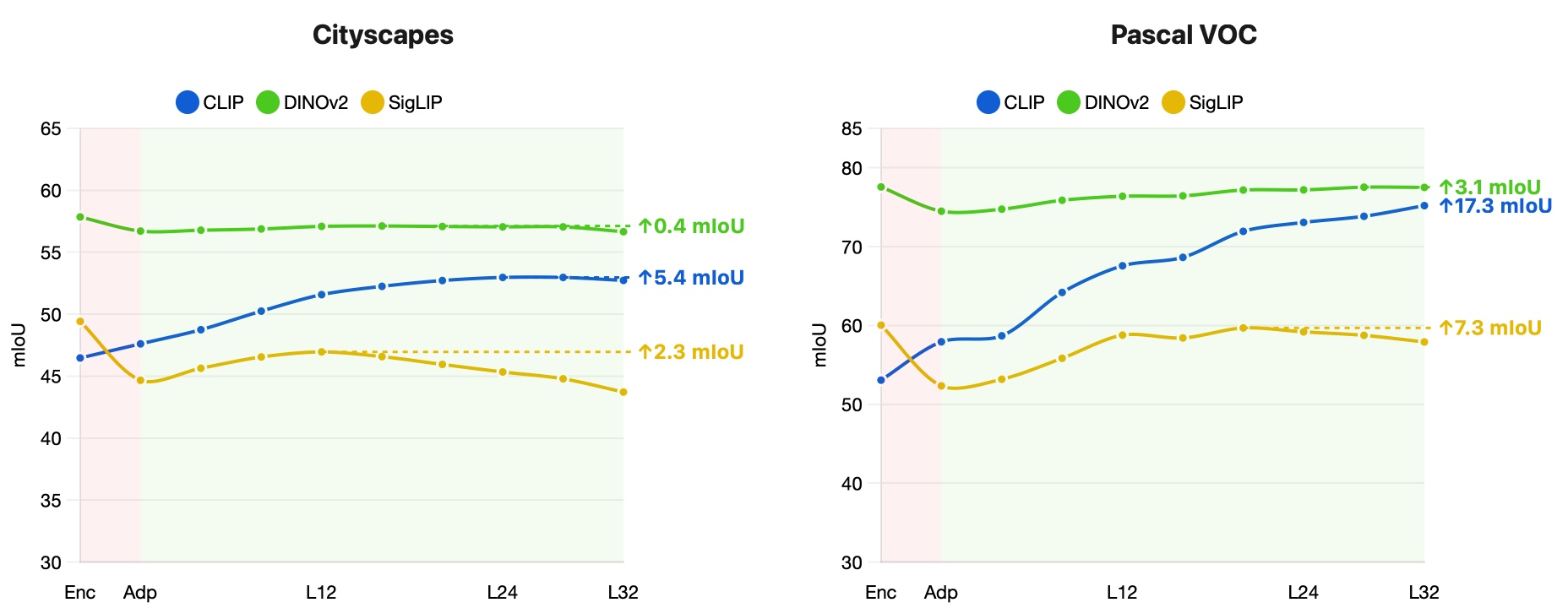}
    \caption{\textbf{Layerwise linear probing on Cityscapes (left) and PASCAL VOC (right).} mIoU across the MLLM stack for CLIP, DINOv2, and SigLIP encoders paired with Vicuna-7B.}
    \label{fig:layerwise-voc-cityscapes}
\end{figure}

The main paper reports layerwise linear probing results on ADE20K (\cref{fig:layerwise-recovery}).
\cref{fig:layerwise-voc-cityscapes} shows the corresponding results on Cityscapes and PASCAL VOC 2012.
The progressive recovery across LLM layers is consistent across all three datasets and all three encoders, confirming that this pattern is not specific to ADE20K.
DINOv2 and SigLIP also exhibit a consistent adapter drop-off across all datasets.

\subsection{Causal vs.\ Bidirectional Probing}

\cref{tab:bidi-probing} summarizes the causal vs.\ bidirectional comparison on ADE20K, complementing the layerwise curves in \cref{fig:bidi_vs_causal_triptych}.

\begin{table}[tb]
\centering
\caption{\textbf{Causal vs.\ bidirectional attention: Layerwise probing on ADE20K.}
Adapter and Peak LLM report mIoU~(\%).
$\Delta_\text{enc}$: Peak LLM mIoU minus vision encoder baseline (positive~$=$ surpasses encoder).
Recovery: Peak LLM mIoU minus adapter mIoU.}
\label{tab:bidi-probing}
\begin{tabular}{@{}llcccc@{}}
\toprule
Encoder & Attention & Adapter & Peak LLM & $\Delta_\text{enc}$ & Recovery \\
\midrule
\multirow{2}{*}{CLIP ViT-L/14@336}
 & Causal        & 33.22 & 40.74 & $+$6.36 & $+$7.52 \\
 & Bidirectional & 32.68 & \textbf{41.16} & $+$6.78 & $+$8.48 \\
\midrule
\multirow{2}{*}{DINOv2 Large}
 & Causal        & 42.90 & 44.50 & $-$1.08 & $+$1.60 \\
 & Bidirectional & 40.47 & \textbf{44.82} & $-$0.76 & $+$4.35 \\
\midrule
\multirow{2}{*}{SigLIP SO400M/14}
 & Causal        & 31.44 & 35.39 & $-$3.31 & $+$3.95 \\
 & Bidirectional & 32.39 & \textbf{41.26} & $+$2.56 & $+$8.87 \\
\bottomrule
\end{tabular}
\end{table}

\subsection{Context Starvation Per-Patch Accuracy}

\cref{tab:context_starvation} reports per-patch pixel accuracy and relative improvement for the first three image tokens under causal and bidirectional attention, quantifying the context starvation effect discussed in \cref{subsec:bidi-results}.

\begin{table}[tb]
  \centering
  \caption{\textbf{Context starvation under causal attention.}
  Per-patch pixel accuracy for the first three tokens.
  $\Delta$: Bidirectional minus causal accuracy. \% Impr.: relative improvement over causal.
  Early patches suffer severe context starvation under causal masking.}
  \label{tab:context_starvation}
  \setlength{\tabcolsep}{10pt}
  \begin{tabular}{@{}l cccc@{}}
    \toprule
    & Causal & Bidirectional & $\Delta$ & \% Impr. \\
    \midrule
    Patch 0 & 0.6195 & 0.7630 & $+$0.1435 & $+$23.2\% \\
    Patch 1 & 0.6851 & 0.7933 & $+$0.1082 & $+$15.8\% \\
    Patch 2 & 0.7069 & 0.7931 & $+$0.0862 & $+$12.2\% \\
    \bottomrule
  \end{tabular}
\end{table}

\subsection{Generalization Across MLLM Architectures}

\begin{figure}[tb]
    \centering
    \includegraphics[width=\linewidth]{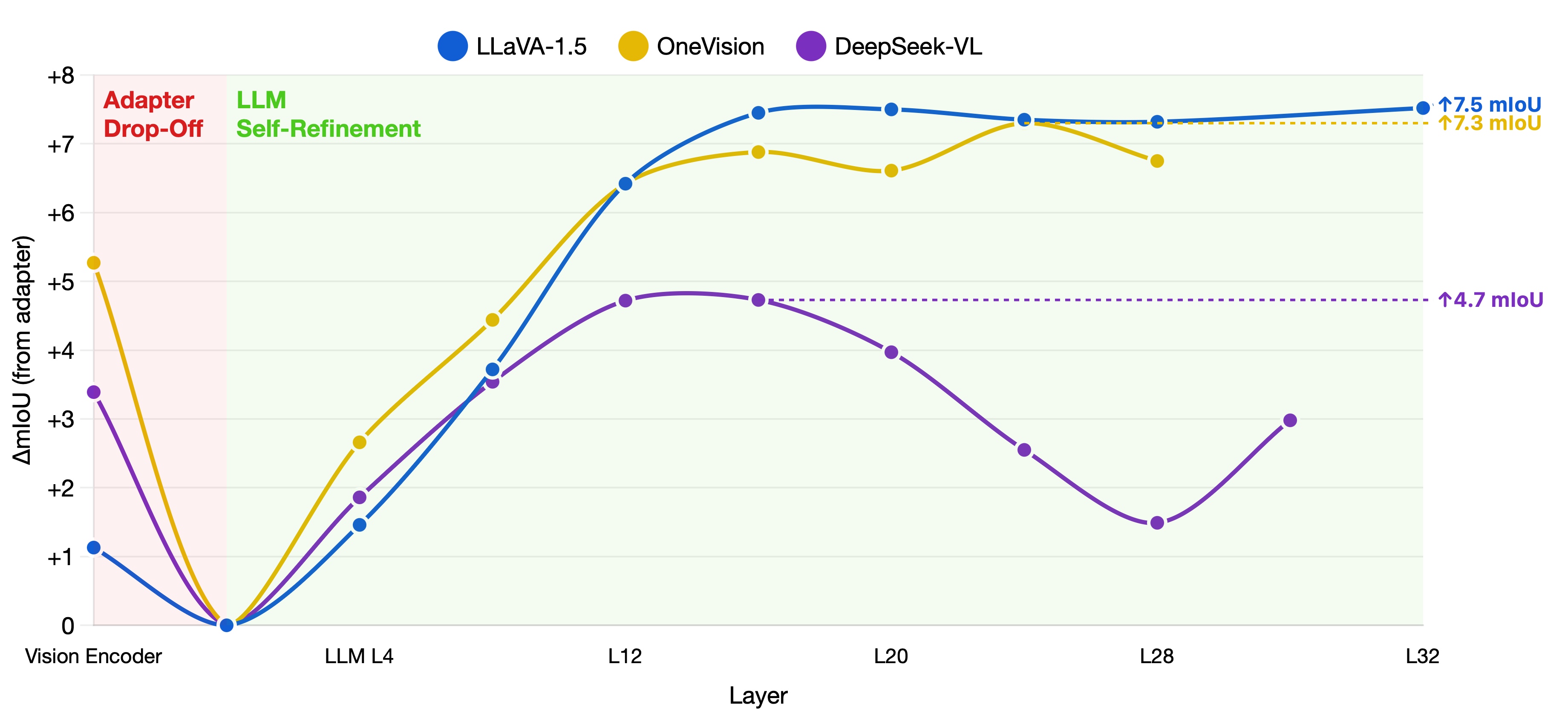}
    \caption{\textbf{Layerwise probing across different MLLMs.} $\Delta$mIoU relative to adapter output on ADE20K for LLaVA-1.5 (7B), LLaVA-OneVision (7B), and DeepSeek-VL (7B), each using its default vision encoder (CLIP for LLaVA-1.5, SigLIP for OneVision, SigLIP+SAM-B for DeepSeek-VL). Curves span different numbers of layers due to differences in LLM backbone depth. The adapter drop-off and LLM self-refinement pattern is consistent across architectures.}
    \label{fig:cross-mllm-recovery}
\end{figure}

\cref{fig:layerwise-recovery} in the main paper varies the vision encoder while keeping the LLM fixed (Vicuna-7B in the LLaVA-1.5 framework).
To test whether the drop-off and recovery pattern generalizes beyond LLaVA-1.5, we repeat the layerwise linear probing experiment with two additional MLLMs: LLaVA-OneVision and DeepSeek-VL. Each model uses a different vision encoder: LLaVA-1.5 uses CLIP ViT-L/14, OneVision uses SigLIP, and DeepSeek-VL uses concatenated SigLIP and SAM-B embeddings.
\cref{fig:cross-mllm-recovery} reports $\Delta$mIoU relative to the adapter output on ADE20K.
All three architectures exhibit the same qualitative pattern: a representation drop-off at the adapter followed by progressive recovery across LLM layers.
Note that the curves span different numbers of layers because each MLLM uses a different LLM backbone with a different depth.
These results confirm that the drop-off and self-refinement mechanism documented in \cref{sec:probing} is not specific to a single MLLM but extends across different adapter-style architectures.

\section{Extended Attention Knockout Experiments}
\label{sec:supp-knockout}

The main paper demonstrates the attention knockout analysis on individual images with a single class-confusion pair (ceiling vs.\ sky).
Here we extend this analysis using the CLIP MLLM to more misclassified classes (e.g. sky, ceiling, and grass) and aggregate the results across many images to provide quantitative evidence that the semantic anchor mechanism generalizes.

\subsection{Aggregate Knockout Metric}

To quantify the knockout effect at scale, we track how many incorrectly predicted patches persist across layers.
The metric below tracks predictions at every layer.
For a given misclassified class $c$ (e.g., sky), we classify all 576 patches at layer~0 using the layer-0 probe and count how many are incorrectly predicted as $c$, yielding $n_{c}^{(0)}$.
At each subsequent layer $\ell$, we count the number of patches still predicted as $c$, yielding $n_{c}^{(\ell)}$.
The rate for that image at layer $\ell$ is $n_{c}^{(\ell)} / n_{c}^{(0)}$.
We average this ratio across all selected images.
A value of 100\% at layer~0 means all initially misclassified patches are still present, lower values indicate the model is correcting them.
We then take the ground-truth class across all misclassified patches in the image as the dominant ground-truth class, this is the class that gets blocked in the knockout-correct condition.

\subsection{Quantitative Results}

\begin{figure}[tb]
    \centering
    \includegraphics[width=0.75\linewidth]{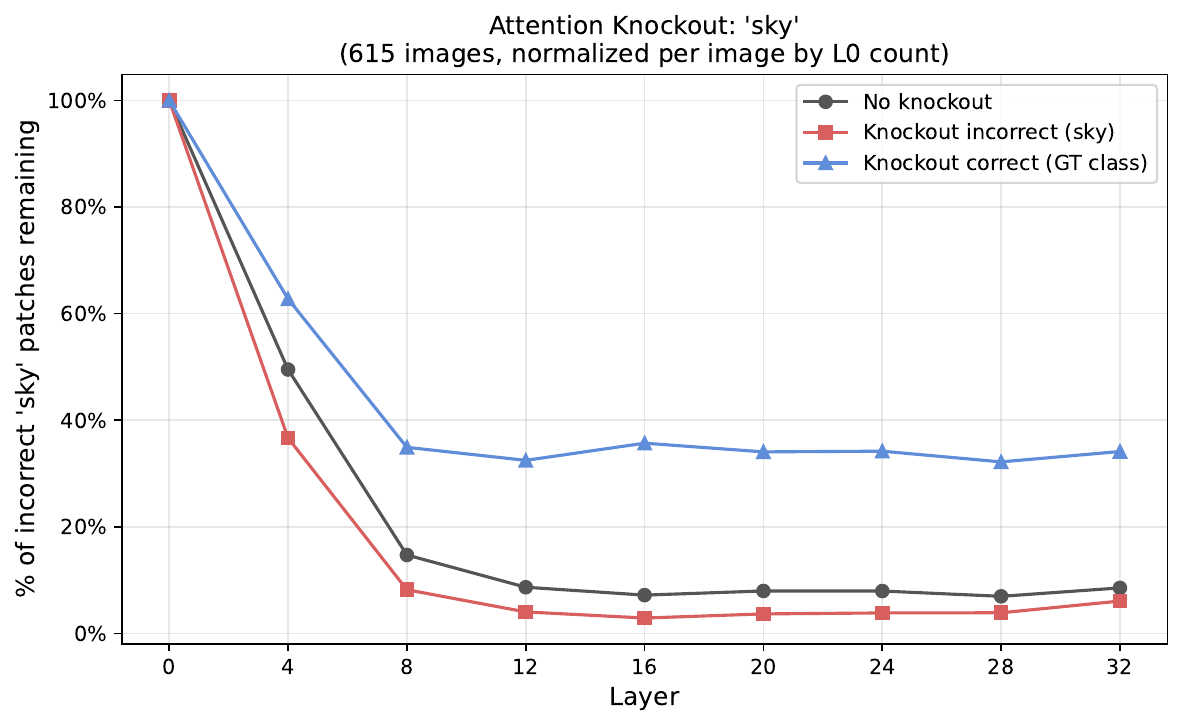}\\[2pt]
    \includegraphics[width=0.75\linewidth]{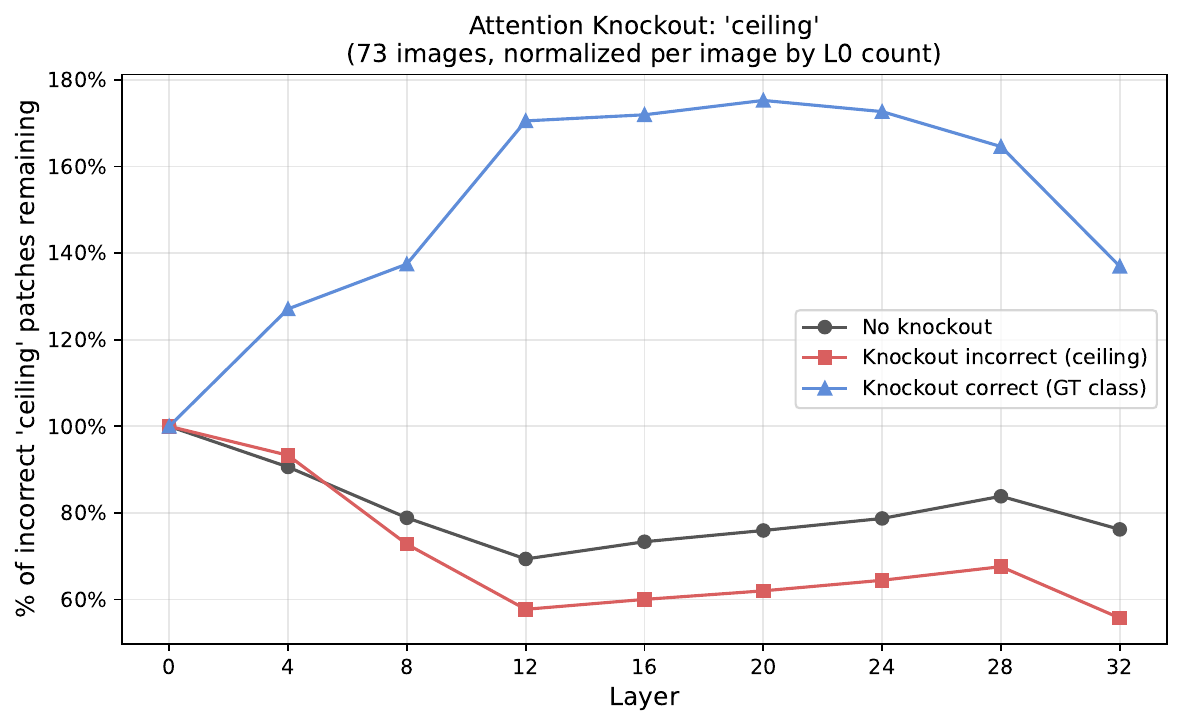}\\[2pt]
    \includegraphics[width=0.75\linewidth]{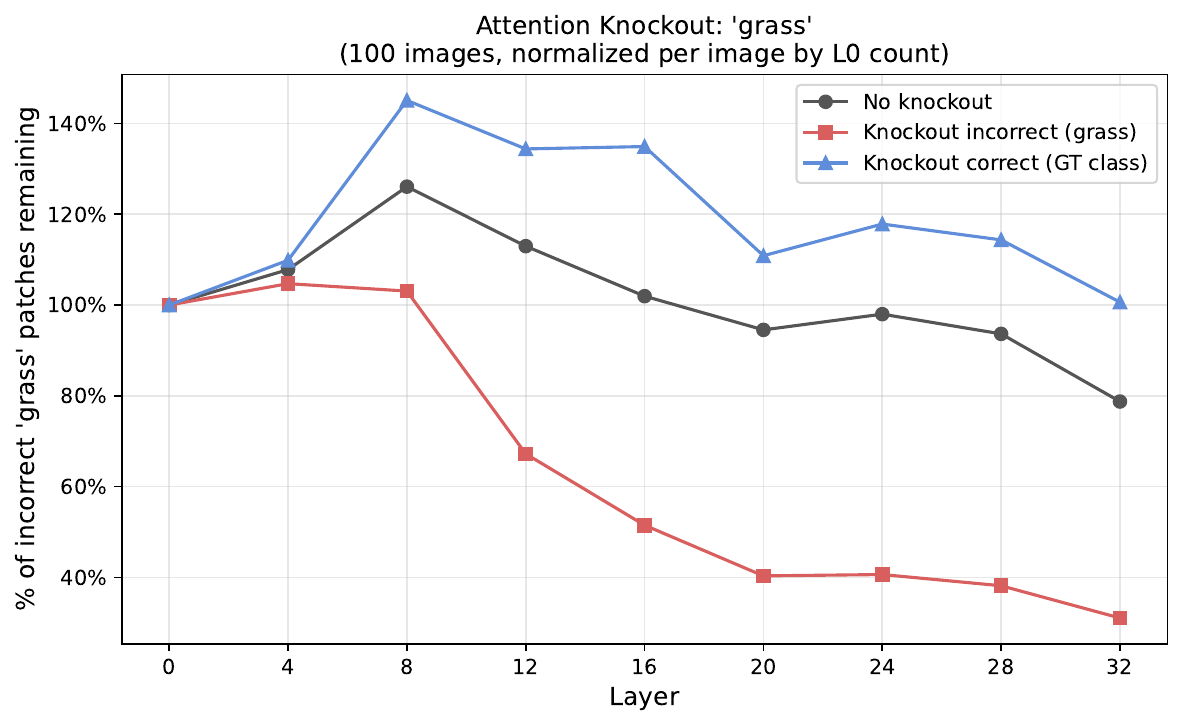}
    \caption{\textbf{Aggregate attention knockout across three misclassified classes.} Percentage of incorrectly predicted patches remaining across LLM layers, normalized to 100\% at layer~0 and averaged over all images exhibiting the misclassification. Blocking the incorrect class (red) accelerates self-correction, blocking the correct class (blue) impairs it and can amplify errors beyond the initial count.}
    \label{fig:knockout-quantitative}
\end{figure}

\cref{fig:knockout-quantitative} reports the aggregate knockout metric for sky, ceiling, and grass.
Across all three classes, blocking the incorrect class accelerates self-correction, while blocking the correct class impairs it.
These results show quantitatively, that correctly classified tokens act as semantic anchors driving self-refinement.

\subsection{Qualitative Examples}

\begin{figure}[tb]
    \centering
    \includegraphics[width=0.65\linewidth]{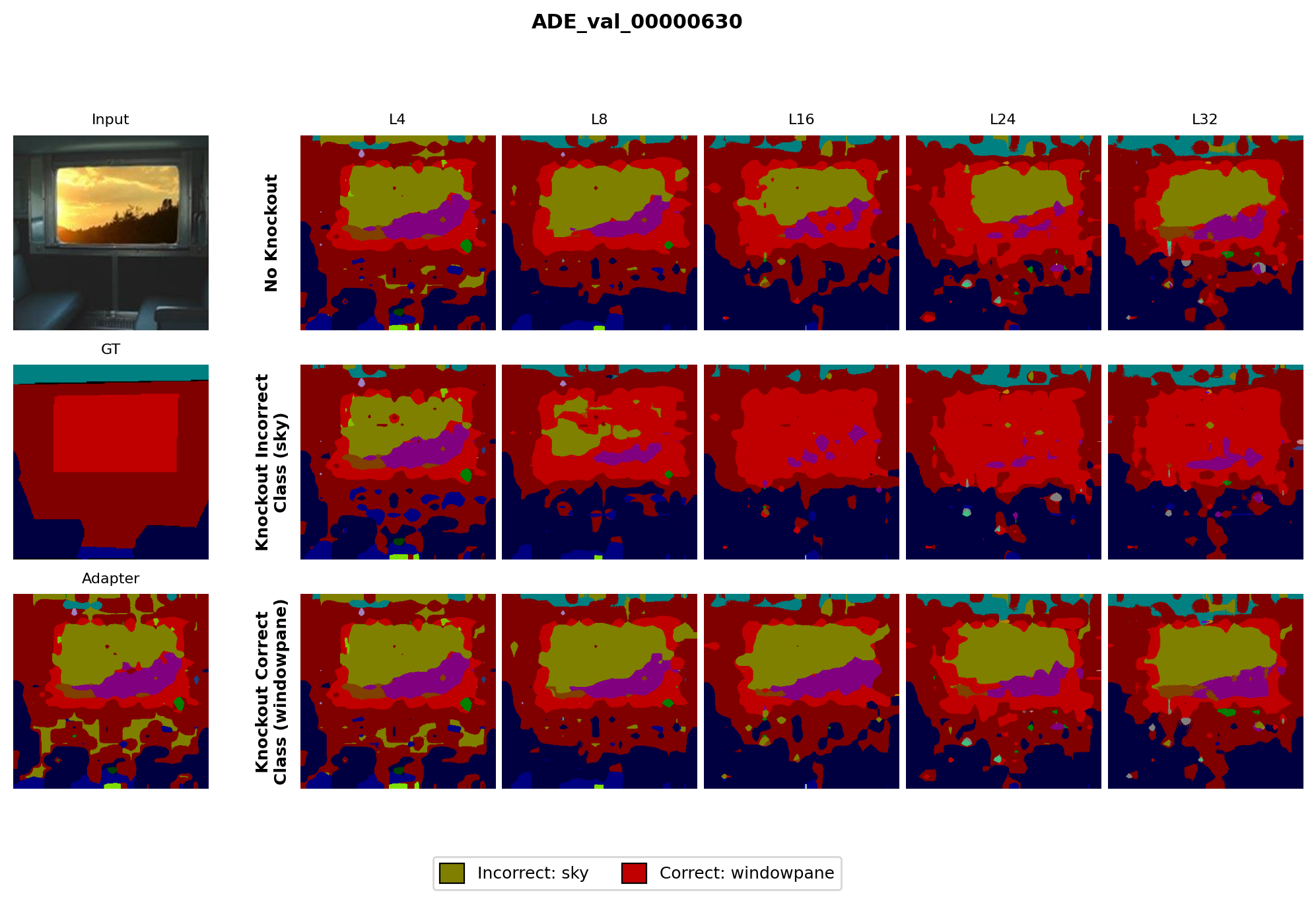}\\[2pt]
    \includegraphics[width=0.65\linewidth]{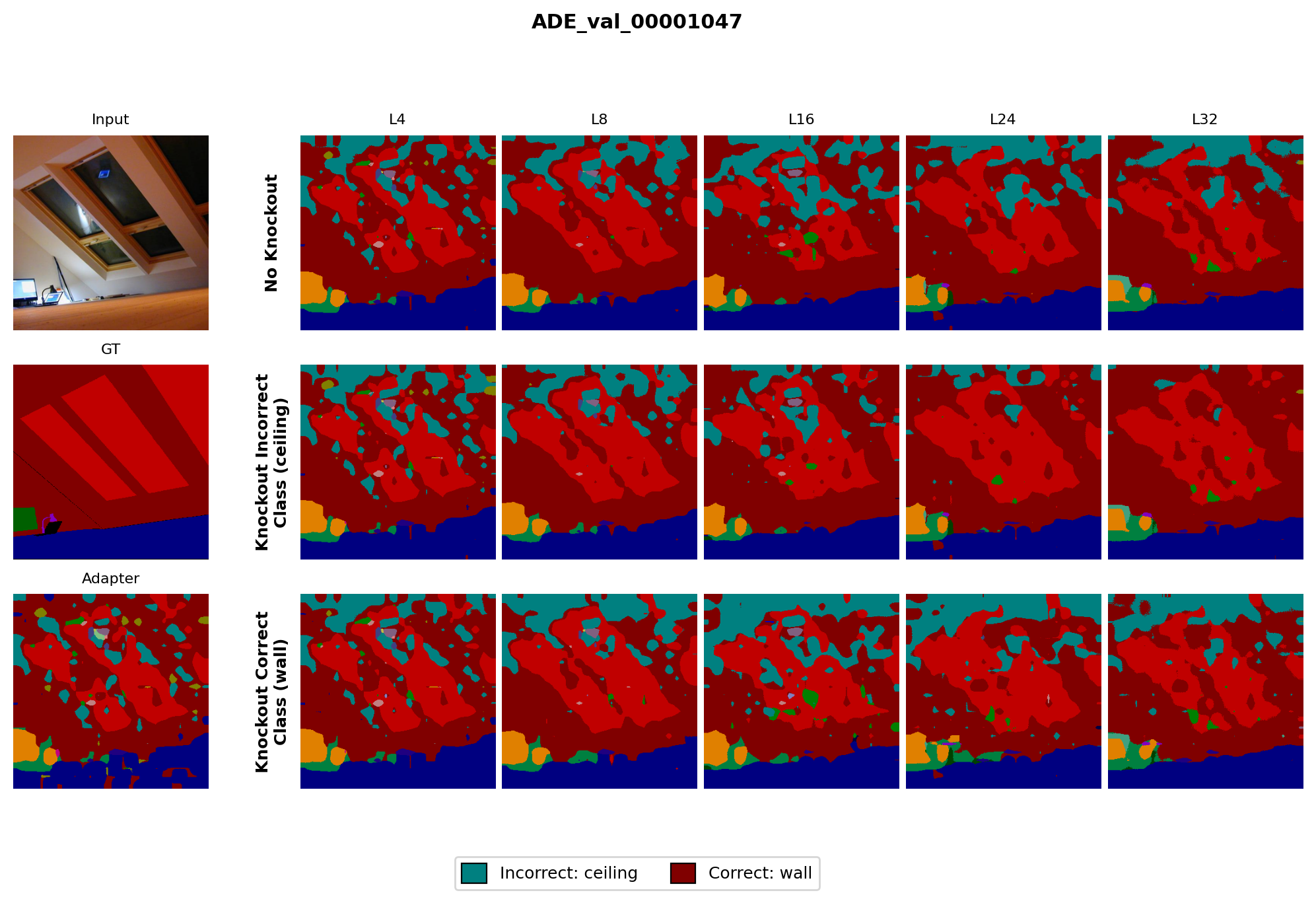}\\[2pt]
    \includegraphics[width=0.65\linewidth]{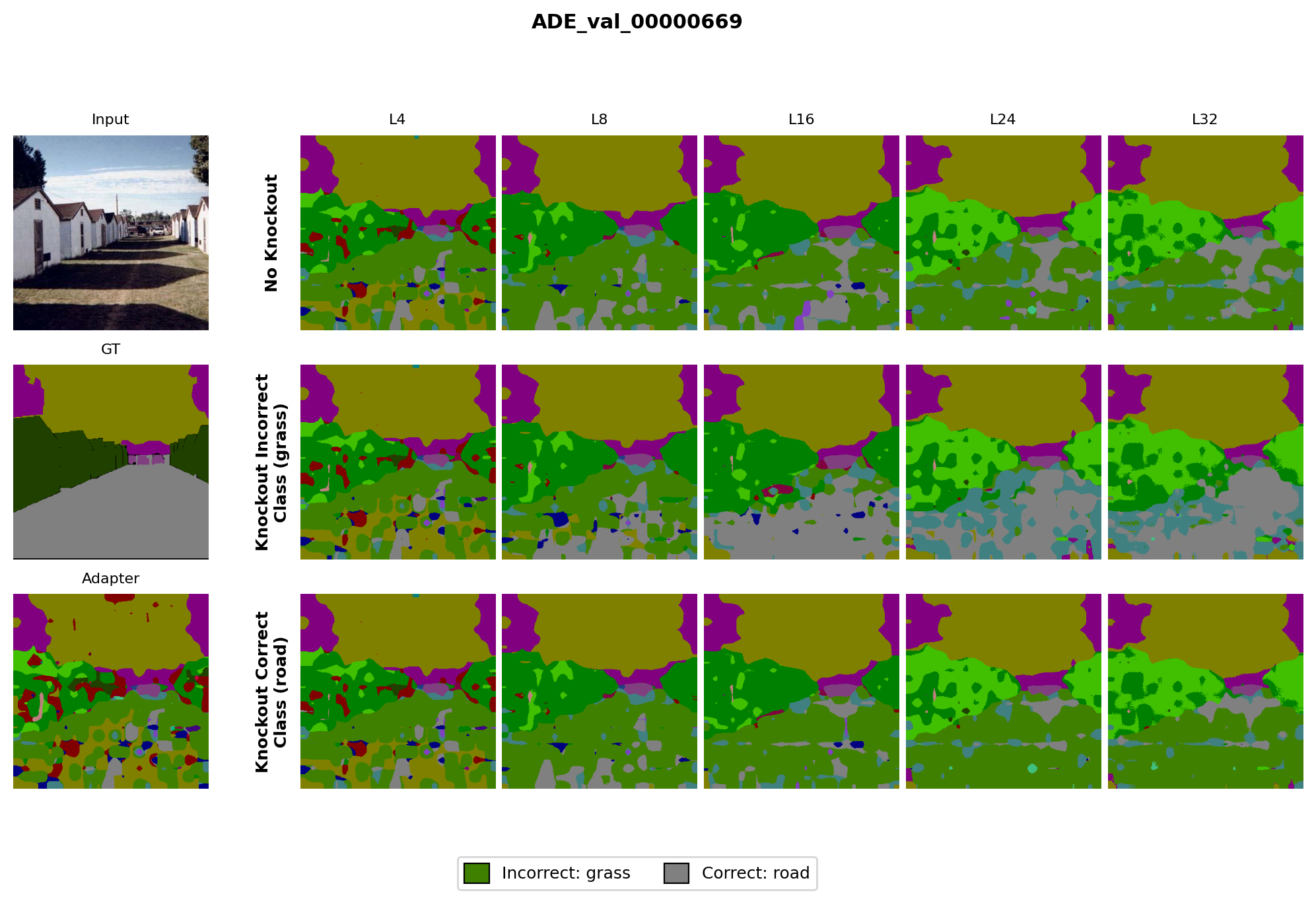}
    \caption{\textbf{Qualitative attention knockout examples.} Layerwise segmentation predictions under three conditions (rows: no knockout, knockout incorrect, knockout correct) for three class-confusion pairs. Top: sky vs.\ windowpane. Middle: ceiling vs.\ wall. Bottom: grass vs.\ road. Blocking the incorrect class produces cleaner segmentation maps at earlier layers, while blocking the correct class leaves errors unresolved or amplified.}
    \label{fig:knockout-qualitative}
\end{figure}

\cref{fig:knockout-qualitative} shows representative per-image knockout visualizations for each of the three misclassified classes, illustrating the layerwise segmentation under all three conditions.

\section{Extended Qualitative Results}
\label{sec:supp-qualitative}

\subsection{Layerwise Segmentation Predictions}

\begin{figure}[tb]
    \centering
    \includegraphics[width=0.75\linewidth]{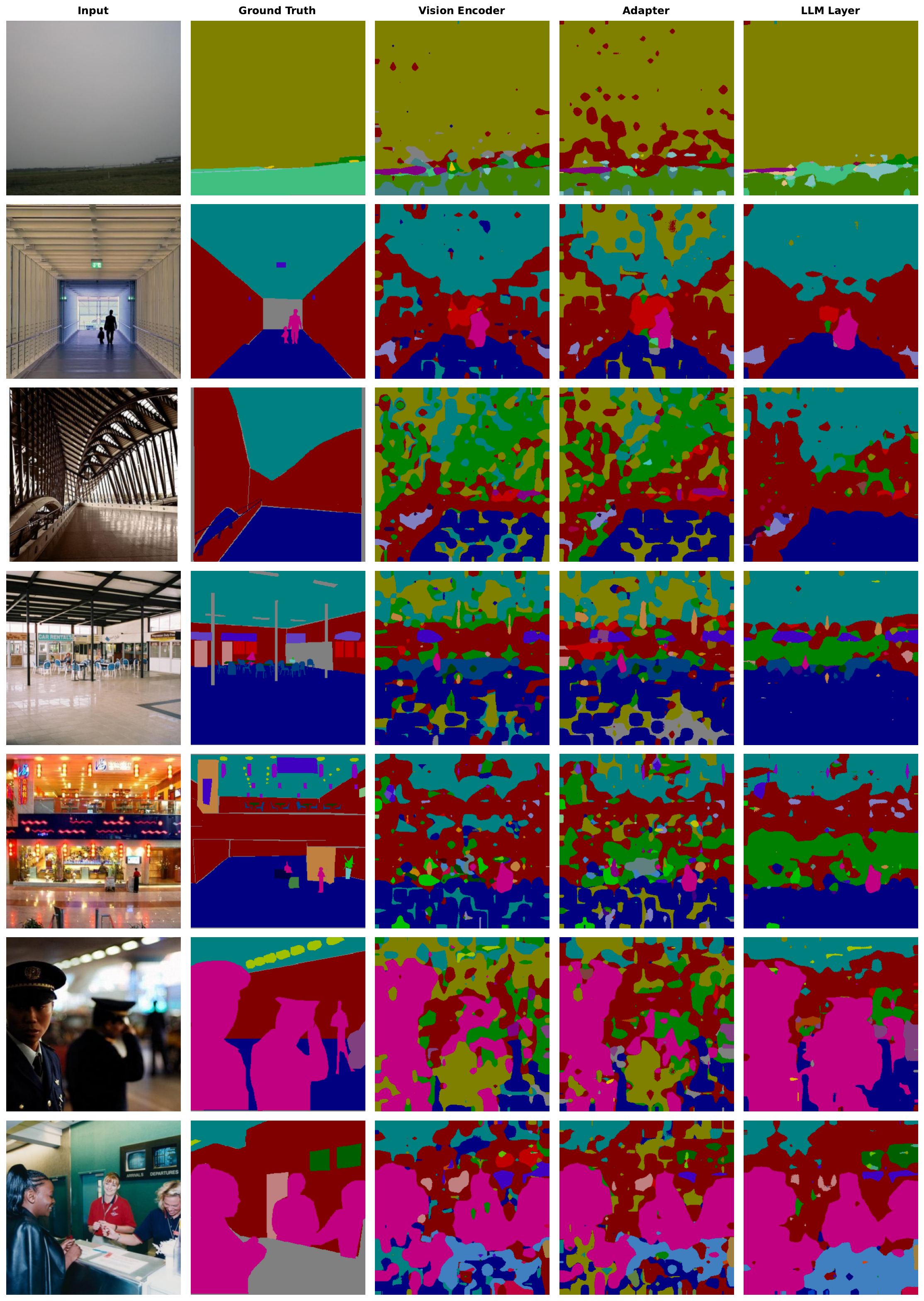}
    \caption{\textbf{Extended qualitative segmentation predictions across the MLLM stack.} Seven ADE20K validation images showing, from left to right: input image, ground truth, linear probe prediction at the vision encoder output, adapter output, and at an intermediate LLM layer. The representation drop-off at the adapter is visible as increased noise and class confusion compared to the vision encoder output, while deeper LLM layers produce more spatially coherent predictions.}
    \label{fig:qualitative-extended}
\end{figure}

\cref{fig:qualitative-extended} shows additional ADE20K validation images with linear probe predictions at the vision encoder output, adapter output and at an intermediate LLM layer, extending the qualitative analysis from \cref{fig:qualitative-probing}.

\end{document}